\documentclass[runningheads]{llncs}
\usepackage[T1]{fontenc}
\usepackage{graphicx}
\usepackage{silence}
\usepackage{graphicx}
\usepackage{booktabs}
\usepackage{lmodern}
\usepackage{siunitx}
\usepackage{etoolbox}
\usepackage{amsfonts}

\usepackage{amsmath}
\usepackage{hyperref}
\usepackage{ifthen}
\usepackage{xcolor}

\usepackage{subcaption}
\usepackage{multirow}
\newboolean{showcomments}
\setboolean{showcomments}{true}
\ifthenelse{\boolean{showcomments}}
{
}

\begin{document}
\title{Monte-Carlo Tree Search for Multi-Agent Pathfinding: Preliminary Results}
\author{
Yelisey Pitanov\inst{1}\and
% \orcidID{0000-0001-9243-1622}
Alexey Skrynnik\inst{2, 3} \and
Anton Andreychuk\inst{2} \and
\\Konstantin Yakovlev\inst{2,3} \and
Aleksandr Panov\inst{2,3}
}
\authorrunning{Y. Pitanov et al.}
\institute{
Moscow Institute of Physics and Technology, Moscow, Russia
% \email{piteles@mail.ru} 
\and
AIRI, Moscow, Russia \and
Federal Research Center "Computer Science and Control" of the Russian Academy of Sciences, Moscow, Russia
}
\maketitle              % typeset the header of the contribution
\begin{abstract}
    In this work we study a well-known and challenging problem of Multi-agent Pathfinding, when a set of agents is confined to a graph, each agent is assigned a unique start and goal vertices and the task is to find a set of collision-free paths (one for each agent) such that each agent reaches its respective goal. We investigate how to utilize Monte-Carlo Tree Search (MCTS) to solve the problem. Although MCTS was shown to demonstrate superior performance in a wide range of problems like playing antagonistic games (e.g. Go, Chess etc.), discovering faster matrix multiplication algorithms etc., its application to the problem at hand was not well studied before. To this end we introduce an original variant of MCTS, tailored to multi-agent pathfinding. The crux of our approach is how the reward, that guides MCTS, is computed. Specifically, we use individual paths to assist the agents with the the goal-reaching behavior, while leaving them freedom to get off the track if it is needed to avoid collisions. We also use a dedicated decomposition technique to reduce the branching factor of the tree search procedure. Empirically we show that the suggested method outperforms the baseline planning algorithm that invokes heuristic search, e.g. A*, at each re-planning step.

\end{abstract}

\section{Introduction}

Pathfinding problems naturally arise in various applications, like autonomous vehicles, household robotics~\cite{noh2022adaptive}, etc. Moreover, in a variety of settings the simultaneous operation of a group of robots is desirable. Thus, multi-agent pathfinding (MAPF)~\cite{stern2019multi} problem arises. Indeed, numerous approaches to this problem are known, both learnable and non-learnable ones, see~\cite{yakovlev2023planning} for an overview. 

In this work we would like to investigate how the machinery of Monte-Carlo Tree Search (MCTS)~\cite{browne2012survey} can be applied to MAPF. The reason is two-fold. First, MCTS, which can be seen as a hybrid approach that integrates planning and learning, has already been successfully applied to various problems with hard combinatorial structure, often outperforming the state-of-the-art competitors. Thus, it is tempting to study how well MAPF problem can be handled by MCTS. The second reason is that if the performance of MCTS is promising than it opens the door to incorporating various deep learning techniques that can further increase its efficiency (similarly to how deep learning was utilized to boost MCTS in playing board games like Go, reaching super-human performance~\cite{silver2017mastering}). In this work we focus on the first part, i.e. how can MCTS be adapted to MAPF and how well it can perform compared to standard search-based approaches.

To this end we introduce an original variant of MCTS that utilizes two key ingredients. The first ingredient is the specific reward-shaping and reward propagating mechanisms that are tailored to MAPF setting. Second, is how the branching factor of the search tree is reduced by decomposing the joint actions into a sequential combination of the individual agent's actions. As confirmed by the numerous experiments, the resultant algorithm is able to successfully solve complex MAPF instances and outperforms a baseline search-based solver.

\section{Background}

\paragraph{Multi-agent pathfinding} in its classical variant~\cite{stern2019multi} assumes that the set of $K$ agents is confined to a graph $G=(V, E)$, where $n$ start and goal vertices are distinguished. The time is discretized and at each time step an agent can wait in its current vertex or move to an adjacent one. The sequence of such actions is called a plan. Two plans for distinct agents are conflict-free if the agents following them never swap vertices at the same time step and never occupy the same vertex at the same time step (i.e. no \emph{edge} and \emph{vertex} collisions are allowed). The task is to find a set of plans $Plans=\{plan_1, plan_2, ..., plan_n\}$, one for each agent, s.t. all agents reach their goals and each pair of plans is conflict-free. Typically, in MAPF one wants to minimize one of the following cost objectives: $SOC=\sum_{i=1}^n cost(plan_i)$ or $makespan = \max_i cost(plan_i)$, where $cost(plan_i)$ is the cost of the individual plan, i.e. the number of time steps it took the agent $i$ to reach its goal.

In this work we are interested in the formulation of MAPF problem as the sequential decision making problem, when at each time step the actions for the agents are decided (based on some action-selection policy), then these actions are executed, and then the cycle repeats until all the agents reach their goals or some predefined threshold on the number of time steps is met. Such framing of the problem is well formalized by Multi-agent Markov Decision Process (MAMDP)~\cite{nawaz2022multi}, which we describe next.

\paragraph{Multi-agent Markov Decision Process (MAMDP)} can formally be represented as \(\langle S, U, T, r, n, \gamma \rangle\). 
Here $n$ is the number of agents and $\textbf{U}=U_1 \times U_2 \times ... \times U_n$ is the joint action that is formed by combining the individual actions of all agents. $S$ is the set of the environment states. In our case each state $s \in S$ encompasses the information on the locations of all agents. 
$T(s'| s, \textbf{u}): S \times \textbf{U} \times S \rightarrow [0, 1]$ is the transition function. It defines the probability of transitioning to state $s'$ when the joint action $\textbf{u} \in \textbf{U}$ is executed at state $s$. $r: S \times \textbf{U} \rightarrow \mathbb{R}$ -- is the reward function shared between all agents. This function specifies what reward (scalar value) the agents will get if they execute a specific action $\textbf{u}$ at the specific environment state $s$. $\gamma \in [0, 1]$ is the discounting factor.

The task is to construct a policy $\pi$ that specifies which actions an agent should take in different states of the environment to maximize the expected return $G$ over an episode with length $K$, as defined by equation \ref{eq:return}:

\begin{equation}\label{eq:return}
G_t = \sum_{k=0}^{K-t-1} \gamma^k r_{t+k+1} \ \textbf{s. t.} \ s_t \in S,
\end{equation}
where $r_{t+k+1}$ is the reward received at time step $t+k+1$, and $s_t$ represents the state of the environment at time step $t$. 
Generally, the policy might be stochastic, i.e. it might map the states to the distribution of actions: $\pi: S \times \textbf{U} \rightarrow [0, 1]$. Given the policy the exact (joint) action is sampled from the distribution at each time step.

Various approaches to obtain $\pi$ for MAPF can be suggested. For example, one can invoke a complete/optimal MAPF solver at each time step and pick the first joint action from the MAPF solution. Another approach may be to invoke $n$ egoistic single agent searches and to construct a joint action by combining the first actions comprising the $n$ single-agents paths that do not take the other agents into account. In this work we will use such an approach as a baseline to compare with. One more direction to look at might be relying on reinforcement learning (RL)~\cite{sutton2018reinforcement} to learn the policy. This, however, might be tricky in the multi-agent setting and non-trivial modifications of the learning-based algorithms are likely to be needed~\cite{rashid2020monotonic}. In this work we, would like, to explore another approach, i.e. to adapt Monte-Carlo Tree Search to multi-agent pathfinding due to the high efficiency of such type of search when it comes to the combinatorial problems with high branching factor.

\paragraph{Monte-Carlo Tree Search (MCTS)} is a powerful search paradigm that is well-suited for the sequential decision making problems. Paired with the state-of-the-art machine learning techniques MCTS has recently achieved super-human performance in various board- and video-games, see~\cite{silver2017mastering,ye2021mastering} for example. As MCTS relies on the notion of reward it can also be attributed to as a model-based reinforcement learning method, that utilizes the reward(s) to learn to pick the promising actions.

In a nutshell MCTS picks an action given a state of the environment based on extensive simulating of how the environment would change and what rewards would be obtained if different (random) actions are sequentially executed. Indeed, it is not possible to simulate all possible sequences of actions in a limited time. To this end MCTS builds a tree (of a limited width anf depth) that contains the most promising candidates of the actions to be taken (partial plans). The nodes in that tree correspond to environment states and edges -- to the actions. The root of the tree is the current state, i.e. the one for which we need to pick an action. 

Particularly, MCTS is composed of the four steps that are executed iteratively and are intended to simultaneously build and explore the search tree: selection, expansion, simulation, and backpropagation. 
Selection is aimed at descending the constructed so far search tree. Conceptually, this can be seen as the process of picking the most promising partial plan to consider. To balance between the exploration (i.e. picking the parts of the search tree that were not considered before) and the exploitation (i.e. picking the partial plans that are characterized by the highest rewards) MCTS relies on assessing the nodes using the upper-confidence bounds techniques (which were initially suggested in the context of the multi-armed bandit problems~\cite{auer2002finite}). Specifically, Upper Confidence Bound for search Trees (UCT)~\cite{kocsis2006bandit} is commonly used. When the tree is descended and the leaf node is picked, the latter is expanded by selecting an un-probed action and adding a new node to the tree. The added node is evaluated by simulating actions using a random policy and the resulting reward is backpropagated through the tree in a special fashion. The process is repeated until the time budget is reached. When it happens the action corresponding to the most visited outgoing edge of the root node is chosen to be executed. For more details on MCTS we refer the reader to an overview paper by Browne et al.~\cite{browne2012survey}. In this work we will present our adaptation of MCTS for the multi-agent pathfinding in the next sections.

\section{Problem statement}
\label{sec:ps}
Consider $n$ homogeneous agents, navigating in the environment which is represented as a 4-connected grid, composed of the free and blocked cells (the latter correspond to the static obstacles of the environment). The timeline is discretized and at each time step an agent can wait at the current cell or move to one of the adjacent cells if it is not blocked. When two agents wish to move to the same free cell only one of them (random one) succeeds, while the other stay where it was. Thus the state transitions are stochastic.

Initially the agents are located at the unique start cells and for each agent a unique goal cell is specified. When an agent reaches the goal it is removed from the grid. This assumption is not uncommon in MAPF as it is reasonable for various practical applications. For example, think of the robots in the warehouse that need to go to the charging stations. Typically, these charging stations are located at the perimeter of the working area, thus a robot that reaches a charging station may be considered to leave the workspace. 

The task is to define a policy that will map environment states to joint actions. Here the environment state is the grid plus the positions of all agents on it. Joint action is a combination of the individual actions of the distinct agents. Moreover, we assume that a limit of $K_{max}$ time steps is given and the episode ends when this number of time steps passes (no matter where the agents are by this time).

To measure how well the policy copes with the problem at hand we use the following metrics, which we compute at the end of the episode:
\begin{itemize}
    \item \emph{Cooperative Success Rate (CSR)} is a boolean metric, i.e. it might be 0 or 1, that indicates whether all the agents manage to reach their goals;
    \item \emph{Individual Success Rate (ISR)} is the fraction of agents that managed to reach their goal before the episode ends.
    \item \emph{Episode Length (EL)} is the number of steps taken by the agents to complete the task. If not all agents reach their goals, EL is assigned the value of $K_{max}$.
\end{itemize}

When comparing different policies the one is better that provides higher CSR/ISR and lower EL. Please note that we do not aim to obtain an optimal policy in this work.

\section{Related Work}
Initially, MCTS algorithms demonstrated their effectiveness in antagonistic games with full information, such as chess or Go \cite{silver2017mastering}. Modern versions of MCTS use deep neural networks to approximate state values instead of simulation. These approaches have also proven effective in single-agent settings where modifications enable agents to learn a model of the environment, allowing them to play Atari games~\cite{schrittwieser2020mastering,ye2021mastering}. MCTS is not tailored only to games, but is used in robotics~\cite{best2019dec,dam2022monte}, theorem proving~\cite{lample2022hypertree} and even can be applied to find an efficient way to multiply matrices~\cite{fawzi2022discovering}. All these examples do, however, belong to the single-agent domain.

Despite the interest in using MCTS for multi-agent settings, a few works have applied it to MAPF. In~\cite{zerbel2019multiagent} the authors propose a multi-agent MCTS for Anonymous MAPF in a grid-world environment. Their environment has a dense reward signal (the agent who reached any goal on the map received a reward and ended the episode), and there were no obstacles present, which simplifies collision avoidance. The authors build a separate tree for each agent using a classical algorithm. They then apply the best actions (forming a plan) from the trees jointly in simulator, to receive true scores of the solution and update the trees on that difference. This approach performs well even with a large number of agents.
A recent paper~\cite{skrynnik2021hybrid} proposed a more sophisticated approach for multi-agent planning that combines RL and MCTS. The authors suggested a two-part scheme that includes a goal achievement module and a conflict resolution module. The latter was trained using MCTS. The construction of the search tree for each of the agents was also performed independently, and actions for other agents were selected using the currently trained policy. In contrast to this approaches, we propose a method which plans over a tree for the whole population of agents in the environment and modifies the reward function to guide the agents towards their goals.

\section{Method}
\label{sec:method}

The straightforward adaptation of MCTS to the considered MAPF problem might be as follows. The nodes in the tree represent the states which are the positions of the agents on a grid, the edges correspond to joint actions. Thus, the branching factor of the tree is $|A|^{n}$, where $|A|$ is the number of possible individual actions and $n$ is the number of agents. For large numbers of agents this leads to great computational overhead. To this end, we suggest to decompose the joint action into the individual ones in the tree, as described in the next sections.

The next problem which arises when one wants to adapt MCTS for MAPF is how to compute the reward at the end of the simulation phase, i.e. after the phase when the agents take random actions. Typically, MCTS is applied to antagonistic games where the outcome is either win (reward equals 1), loss (reward equals 0) or draw (reward equals $1/2$). Thus, one may come with the following reward for our case: $R=n_{finished}/n$, where $n_{finished}$ is the number of agents that reached their goals. However, such a reward is extremely sparse, i.e. in numerous simulation rollouts the reward will be zero or very close to zero (as it is very hard to reach the goal by randomly moving on a grid for a limited number of steps). Thus it will be very hard to focus the search on promising parts of the tree. To this end we introduce an auxiliary intrinsic reward for every agent based on how often it reaches the cells lying on the shortest (individual) path to its goal. Intuitively, such reward shaping forces the agents to demonstrate the goal-reaching behavior while leaving them the freedom to step off the path when needed, e.g. when some agents need to give way to the other agents.

In the next sections we will elaborate more on how our variant of MCTS is designed.

\subsection{Subgoal Based Reward Function}

MCTS is effective in antagonistic games because they  typically have a limited number of moves and a definite outcome (loss, win, draw). In MAPF, however, episodes may last longer and often the randomly-moving agents do not reach their goals. In such simulations, the agents may not receive a positive reward signal to guide the search.

We propose a technique that addresses this issue by encouraging subgoal achievement through intrinsic rewards, using a classic planning algorithm such as A*. The planning algorithm is used to find a path, ignoring other agents. The chosen subgoal is placed a short distance from the agent, for example, two steps away. 
The reward function is presented in equation~\ref{eq:reward}:

\begin{equation}\label{eq:reward}
\begin{aligned}
    r_t = \sum_{i=0}^{n-1} \frac{r^{a_i}_t}{n}, &\qquad
 r^{a_i}_t = \begin{cases}
r_{target}, & \text{if the agent}\ a_i\ \text{reached the target,} \\
r_{subgoal}, & \text{if the agent}\ a_i\ \text{reached the subgoal,} \\
0, & \text{otherwise.}
\end{cases}
\end{aligned}
\end{equation}
Achieving the subgoal yields a small positive reward (denoted by $r^s \in R$) that is less than the global target reward (denoted by $r^g \in R$) such that $r^s \ll r^g$. The reward comes from the environment for all agents when performing an joint action. This reward is divided by the number of agents in the environment to ensure consistency with the UCT exploration bonus.
If the agent moves too far from the subgoal, the subgoal position is recalculated.

\subsection{Multi-agent MCTS}

\begin{figure}[ht!]
    \vspace{-15px}
    \centering
    \includegraphics[width=1.0\textwidth]{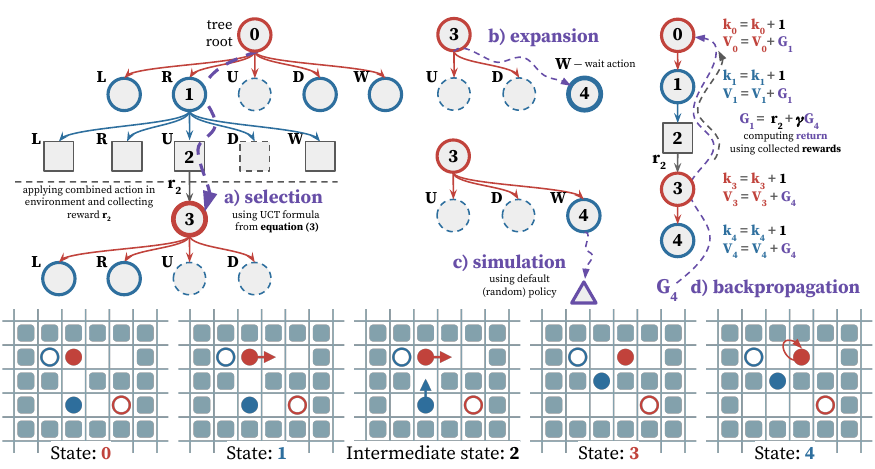}
    \caption{The MAMCTS approach adapts MCTS for multi-agent tasks by treating action selection points for each agent as separate nodes in the tree, thereby reducing the branching factor. The figure shows all four phases of MCTS: a) selection; b) expansion; c) simulation; and d) backpropagation for multi-agent settings. Actions leading to the vertices shown with dotted lines are not considered during the selection phase. Filled circles represent agents, while empty circles represent their corresponding targets. }
    \label{fig:mamcts}
    \vspace{-10px}
\end{figure}

Naive application of MCTS to MAPF assumes generating $|A|^{n}$ child nodes for a search state (leave in the tree), where $|A|$ is the number of individual actions and $n$ is the number of agents. Indeed, this is impractical. To reduce the branching factor we suggest a decomposition approach outlined in Figure~\ref{fig:mamcts} (we note that this resembles the method suggested in~\cite{standley2010finding}). We treat action selection points for each agent as separate nodes in the tree. Actions accumulated in this way are applied at the moment when the last agent selects their action (these nodes are represented as squares in the figure). Thus, the state of the environment is determined by the current positions of the agents and the accumulated actions of other agents. If the accumulated actions are insufficient to perform a joint action to start simulation, the default policy selects the remaining action. Additionally, the tree restricts the choice of actions that lead to static obstacles.

Node selection is performed using the UCT exploration bonus $UCT = \frac{V_j}{k_j} + C_p \sqrt{\frac{2*ln(k)}{k_j}}$
, like in classical MCTS, but now for each separate node. In this equation, $V_j$ is the total accumulated return of the j-th child node, $C_p$ is the exploration coefficient, $k$ is the number of visits to the parent node, and $k_j$ is the number of visits to the child node. 

The sequential approach to action selection also changes the calculation of the return $G$ (see equation~\ref{eq:return}), now the discounting occurs only once between the execution of joint actions in the environment (between the square vertices in the figure). The vertices are updated using the same reward function value after accumulating the overall action. The return $G$ is obtained during simulation with default (random) policy is computed in a standard way. The RL-style value function (with discounting, i.e. $\gamma < 1$) is used to prioritize shorter paths.

\section{Experimental Setup}

\paragraph{Environment.} We used \textsc{POGEMA}\footnote{\href{https://github.com/AIRI-Institute/pogema}{https://github.com/AIRI-Institute/pogema}}~\cite{skrynnik2022pogema,Skrynnik2022a}, an open-source multi-agent pathfinding environment, for our experiments. Initially designed for partial observability problems, it worked well for our setting of full observability. Instead of pickling states for later restoration, we implemented a rollback action that restored the environment to its original state, which is necessary for tree planning. This approach yielded better computational performance than naive state pickling.

\begin{figure}
    \vspace{-10px}
    \centering

    \begin{subfigure}[b]{0.48\textwidth}
            \centering
            \includegraphics[width=\linewidth]{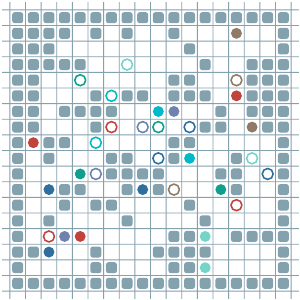}
            \caption{Cooperative random map}
             \label{sfig:a}
    \end{subfigure}
    \hfill
    \begin{subfigure}[b]{0.48\textwidth}
            \centering
            \includegraphics[width=\linewidth]{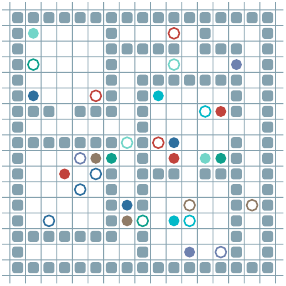}
            \caption{Labmaze map}
             \label{sfig:b}
    \end{subfigure}
    \caption{Examples of maps used during training: (a) Cooperative random maps, which were selected according to their difficulty. (b) Maze maps generated with the \textsc{labmaze} package. Agents are represented by filled circles and their targets are represented by empty circles. Each agent has its own target.}
    \label{fig:map-examples}
    \vspace{-20px}
\end{figure}

\paragraph{Maps.} We used two types of maps for evaluation. The first class of maps includes grid with randomly blocked cells. While random maps are good for their diversity, they do not pose complex challenges to the algorithm because narrow passages that require cooperative behavior rarely appear on the map. To demonstrate the potential of cooperative behavior, we used an approach proposed in~\cite{skrynnik2021hybrid} and created a set of challenging maps sorted by the intersection of individual paths of $16$ agents. We used $2000$ seeds when selecting maps and chose the $100$ most challenging ones with sizes $16\times16$. These maps typically have only one central passage, forcing agents to act jointly. We report the example of a such cooperative map in Figure~\ref{fig:map-examples} (a). To generate them, we used an approach embedded in the \textsc{POGEMA} environment, with an obstacle density of $0.3$. We also conducted post-processing on this set by filling empty components with obstacles that were smaller than five cells.

Secondly, we used maze maps generated with the \textsc{labmaze} package\footnote{\href{https://github.com/deepmind/labmaze}{https://github.com/deepmind/labmaze}}, that contains patterns of mazes and rooms with multiple entrances. These maps include many narrow passages and by design are more challenging than the random maps. Thus, we did not specifically select difficult seeds for this set. We beleive that the ``difficulty'' of these maps is approximately the same as that of cooperative random maps. We used a size of $15\times15$ when generating the maps (since the \textsc{labmaze} package only supports odd sizes).  The example of the generated map is presented in Figure~\ref{fig:map-examples} (b). 
The parameters used for generating the labmaze set of maps were: 
max\_rooms: $30$, has\_doors: True, room\_min\_size: 
$5$, room\_max\_size: $5$, simplify: False, min\_component\_size: $4$, retry\_count: $1000$, extra\_connection\_probability: $0.0$.

\paragraph{Algorithms.} We were comparing four algorithms: Joint MCTS, MAMCTS, Subgoal MAMCTS, and modified A*~\cite{ju2020path}. Joint MCTS is a naive variant of MCTS, where action selection in the tree occurs in a joint action space of all agents. The second approach is MAMCTS, where the action space is reduced and only one agent is considered at each step of tree node expansion. The third variant is Subgoal MAMCTS, which uses a modified reward function that includes a signal for achieving subgoals (as described in Section~\ref{sec:method}). The A* algorithm was originally used for heuristic path planning in single-agent tasks. The algorithm constructs a plan on the full map, considering other agents as obstacles. The first action from the algorithm's plan is selected as the action in the environment. Preliminary experiments have shown that this straightforward approach works well for tasks with a small number of agents and maps without narrow passages. However, if agents need to interact, for instance, one agent has to let another pass, then the algorithm encounters significant issues. To overcome these issues, we introduced a heuristic -- if the last two actions of the algorithm do not lead to the target, the agent chooses a random action instead.

During testing, all MCTS-based approaches received a reward of $r_{target}=+1$ for each agent achieving the goal (notably, the A* approach doesn't exploit this information). The episode of interaction with the environment lasted for $64$ steps. The random priority approach was used as a conflict resolution system. If multiple agents attempted to move to the same cell at the same time step, one agent was randomly selected to occupy the cell, while the others remained in their original positions.  When constructing the tree, it was forbidden to choose actions that lead to static obstacles. During execution MCTS algorithms performed $1000$ selection-expansion iterations, using exploration coefficient $C_p=1$. A discount factor of $\gamma=0.9$ was used.

The Subgoal MAMCTS algorithm provides a reward signal of $r_{subgoal}=+0.1$ for reaching a subgoal, which is the cell located two steps away from the agent along the shortest path (considering cells occupied by other agents as traversable). The subgoal is recalculated each time the agent reaches it or moves 2 steps away from it. This distance was determined experimentally in a preliminary experiment. We also used a limit of $10$ simulation steps for this algorithm, which speeds it up considerably without compromising the results. For other algorithms, this restriction was not applied, because it significantly worsened their results.

\section{Experimental Results}

Running the algorithms on the coperative random maps lead to the results presented in Table~\ref{tab:exp-random-maps}.
Here, for each algorithm, the averaged values of the metrics introduced in Section~\ref{sec:ps} are provided. First, it is worth noting that MAMCTS outperforms Joint MCTS. The action space of the Joint MCTS algorithm is too large, since exponentially dependent on the number of agents in the environment. Planning just one step ahead in the tree requires calculating results for $5^n$ possible nodes, significantly slowing down the search process. The difference is especially noticeable on maps with a large number of agents. The reward signal when an agent reaches the finish line is also quite rare, as can be seen by comparing the results of the MAMCTS and Subgoal MAMCTS algorithms. An additional reward signal solves the problem of long-term planning.

\begin{table}[ht!]
    \vspace{-15px}
    \centering
    \caption{The performance of different algorithms was compared on 100 cooperative random maps of size 16x16. }
    \label{tab:exp-random-maps}
    \vspace{4px}
    \resizebox{\textwidth}{!}{%
                \begin{tabular}{c c c c c c c c c c c c c}
                    \toprule
                    & \multicolumn{3}{c}{\textsc{Joint MCTS}} & \multicolumn{3}{c}{\textsc{MAMCTS}} & \multicolumn{3}{c}{\textsc{Subgoal MAMCTS}} & \multicolumn{3}{c}{\textsc{A*}} \\
                    \cmidrule(lr){2-4}\cmidrule(lr){5-7}\cmidrule(lr){8-10}\cmidrule(lr){11-13}
                    Agents & ISR & CSR & Ep. len. & ISR & CSR & Ep. len. & ISR & CSR & Ep. len. & ISR & CSR & Ep. len. \\
                    \midrule
                    4 & 0.31 & 0.04 & 49.37 & 0.43 & 0.0 & 41.31 & \textbf{1.0} & \textbf{1.0} & \textbf{18.8} & 0.84 & 0.75 & 24.73  \\
                    8 & 0.29 & 0.0 & 50.83 & 0.48 & 0.0 & 40.08 & \textbf{0.99} & \textbf{0.93} & \textbf{21.7} & 0.66 & 0.39 & 32.3  \\
                    16 & 0.24 & 0.0 & 53.3 & 0.4 & 0.0 & 44.47 & \textbf{0.9} & \textbf{0.21} & \textbf{30.14} & 0.4 & 0.0 & 44.21    \\
                    \bottomrule
                \end{tabular}}
    \vspace{-10px}
\end{table}

The modification of the A* algorithm performs well and outperforms all tree search algorithms except for Subgoal MAMCTS. The difference is especially noticeable on maps with $4$ and $8$ agents. Conflicts rarely occur in such cases, and the algorithm is sufficient even with random actions to resolve them.

The comparison of algorithms on maze maps largely repeats the results of the previous experiment. The results a reported in Table~\ref{tab:maze-maps}. The Subgoal MAMCTS approach remains the leader among the considered algorithms, followed by a modified A* algorithm in second place. It is also noteworthy that this set of maps turned out to be slightly easier than cooperative random maps.

Additionally, we compared the computational efficiency of the algorithms. I.e. we computed how long did it take  for an algorithm to choose the actions for 16 agents (on average across all types of maps). On average, Joint MCTS needs $12.1$ seconds to decide actions, MAMCTS takes $12.7$ seconds, and Subgoal MAMCTS takes $4.2$ seconds due to the limited number of steps in the simulation. As expected, the A* algorithm performs the best, taking only $0.02$ seconds to choose actions.   

\begin{table}[ht!]
    \vspace{-10px}
    \centering
    % \scriptsize
    \caption{Testing results of algorithms on $100$ labmaze maps of size $15\times15$. The Subgoal MAMCTS algorithm showed the best results.}
    \label{tab:maze-maps}
    \vspace{4px}
    \resizebox{\textwidth}{!}{%
        \begin{tabular}{c c c c c c c c c c c c c}
            \toprule
            & \multicolumn{3}{c}{\textsc{Joint MCTS}} & \multicolumn{3}{c}{\textsc{MAMCTS}} & \multicolumn{3}{c}{\textsc{Subgoal MAMCTS}} & \multicolumn{3}{c}{\textsc{A*}} \\
            \cmidrule(lr){2-4}\cmidrule(lr){5-7}\cmidrule(lr){8-10}\cmidrule(lr){11-13}
            Agents & ISR  & CSR  & Ep. len.  & ISR & CSR & Ep. len. & ISR & CSR & Ep. len. & ISR & CSR & Ep. len. \\
            \midrule
            4 & 0.39 & 0.0 & 48.25 & 0.54 & 0.0 & 37.74 & \textbf{1.0} & \textbf{1.0} & \textbf{17.17} & 0.9 & 0.79 & 22.41  \\
            8 & 0.32 & 0.0 & 50.82 & 0.55 & 0.0 & 39.16 & \textbf{0.98} & \textbf{0.89} & \textbf{20.18} & 0.83 & 0.54 & 26.42  \\
            16 & 0.22 & 0.0 & 54.19 & 0.5 & 0.0 & 42.02 & \textbf{0.94} & \textbf{0.46} & \textbf{25.64} & 0.58 & 0.07 & 38.62    \\
            \bottomrule
        \end{tabular}}
    \vspace{-15px}
\end{table}

In conclusion, it can be inferred that the proposed algorithm, Subgoal MAMCTS, showed better results compared to all other algorithms. Reducing the branching factor by dividing the action space among agents, as well as adding an additional reward signal that guides the agent towards the target, were key factors in the success of this algorithm. On the downside, our current implementation of Subgoal MCTS is evidently slower compared to the search-based baseline.

\section{Conclusion}

In this study, we applied MCTS to the multi-agent pathfinding problem and proposed techniques to enhance its performance. Our MAMCTS outperforms vanilla MCTS and simple planning algorithms on various map variants, particularly in scenarios with high agent density. Although MAMCTS may not be as fast as individual pathfinding policies based on A* search, it proves efficient given sufficient computational resources and time. In the future, we plan to incorporate neural networks to accelerate the algorithm by approximating the policy and value function. Additionally, applying MCTS to the multi-agent problem opens up possibilities for solving more complex environments with changing map topology and agents possessing extra actions.

\bibliographystyle{splncs}
\bibliography{bib}

\begin{thebibliography}{10}
\providecommand{\url}[1]{\texttt{#1}}
\providecommand{\urlprefix}{URL }
\providecommand{\doi}[1]{https://doi.org/#1}

\bibitem{auer2002finite}
Auer, P., Cesa-Bianchi, N., Fischer, P.: Finite-time analysis of the multiarmed
  bandit problem. Machine learning  \textbf{47},  235--256 (2002)

\bibitem{best2019dec}
Best, G., Cliff, O.M., Patten, T., Mettu, R.R., Fitch, R.: Dec-mcts:
  Decentralized planning for multi-robot active perception. The International
  Journal of Robotics Research  \textbf{38}(2-3),  316--337 (2019)

\bibitem{browne2012survey}
Browne, C.B., Powley, E., Whitehouse, D., Lucas, S.M., Cowling, P.I.,
  Rohlfshagen, P., Tavener, S., Perez, D., Samothrakis, S., Colton, S.: A
  survey of monte carlo tree search methods. IEEE Transactions on Computational
  Intelligence and AI in games  \textbf{4}(1),  1--43 (2012)

\bibitem{dam2022monte}
Dam, T., Chalvatzaki, G., Peters, J., Pajarinen, J.: Monte-carlo robot path
  planning. IEEE Robotics and Automation Letters  \textbf{7}(4),  11213--11220
  (2022)

\bibitem{fawzi2022discovering}
Fawzi, A., Balog, M., Huang, A., Hubert, T., Romera-Paredes, B., Barekatain,
  M., Novikov, A., R~Ruiz, F.J., Schrittwieser, J., Swirszcz, G., et~al.:
  Discovering faster matrix multiplication algorithms with reinforcement
  learning. Nature  \textbf{610}(7930),  47--53 (2022)

\bibitem{ju2020path}
Ju, C., Luo, Q., Yan, X.: Path planning using an improved a-star algorithm. In:
  2020 11th International Conference on Prognostics and System Health
  Management (PHM-2020 Jinan). pp. 23--26. IEEE (2020)

\bibitem{kocsis2006bandit}
Kocsis, L., Szepesv{\'a}ri, C.: Bandit based monte-carlo planning. In: Machine
  Learning: ECML 2006: 17th European Conference on Machine Learning Berlin,
  Germany, September 18-22, 2006 Proceedings 17. pp. 282--293. Springer (2006)

\bibitem{lample2022hypertree}
Lample, G., Lacroix, T., Lachaux, M.A., Rodriguez, A., Hayat, A., Lavril, T.,
  Ebner, G., Martinet, X.: Hypertree proof search for neural theorem proving.
  Advances in Neural Information Processing Systems  \textbf{35},  26337--26349
  (2022)

\bibitem{nawaz2022multi}
Nawaz, F., Ornik, M.: Multi-agent multi-target path planning in markov decision
  processes. arXiv preprint arXiv:2205.15841  (2022)

\bibitem{noh2022adaptive}
Noh, D., Lee, W., Kim, H.R., Cho, I.S., Shim, I.B., Baek, S.: Adaptive coverage
  path planning policy for a cleaning robot with deep reinforcement learning.
  In: 2022 IEEE International Conference on Consumer Electronics (ICCE).
  pp.~1--6. IEEE (2022)

\bibitem{rashid2020monotonic}
Rashid, T., Samvelyan, M., De~Witt, C.S., Farquhar, G., Foerster, J., Whiteson,
  S.: Monotonic value function factorisation for deep multi-agent reinforcement
  learning. The Journal of Machine Learning Research  \textbf{21}(1),
  7234--7284 (2020)

\bibitem{schrittwieser2020mastering}
Schrittwieser, J., Antonoglou, I., Hubert, T., Simonyan, K., Sifre, L.,
  Schmitt, S., Guez, A., Lockhart, E., Hassabis, D., Graepel, T., et~al.:
  Mastering atari, go, chess and shogi by planning with a learned model. Nature
   \textbf{588}(7839),  604--609 (2020)

\bibitem{silver2017mastering}
Silver, D., Schrittwieser, J., Simonyan, K., Antonoglou, I., Huang, A., Guez,
  A., Hubert, T., Baker, L., Lai, M., Bolton, A., et~al.: Mastering the game of
  go without human knowledge. nature  \textbf{550}(7676),  354--359 (2017)

\bibitem{Skrynnik2022a}
Skrynnik, A., Andreychuk, A., Yakovlev, K., Panov, A.: Pathfinding in
  stochastic environments: learning vs planning. PeerJ Computer Science
  \textbf{8},  e1056 (2022). \doi{10.7717/peerj-cs.1056},
  \url{https://peerj.com/articles/cs-1056}

\bibitem{skrynnik2022pogema}
Skrynnik, A., Andreychuk, A., Yakovlev, K., Panov, A.I.: Pogema: partially
  observable grid environment for multiple agents. arXiv preprint
  arXiv:2206.10944  (2022)

\bibitem{skrynnik2021hybrid}
Skrynnik, A., Yakovleva, A., Davydov, V., Yakovlev, K., Panov, A.I.: Hybrid
  policy learning for multi-agent pathfinding. IEEE Access  \textbf{9},
  126034--126047 (2021)

\bibitem{standley2010finding}
Standley, T.: Finding optimal solutions to cooperative pathfinding problems.
  In: Proceedings of the AAAI Conference on Artificial Intelligence. vol.~24,
  pp. 173--178 (2010)

\bibitem{stern2019multi}
Stern, R., Sturtevant, N., Felner, A., Koenig, S., Ma, H., Walker, T., Li, J.,
  Atzmon, D., Cohen, L., Kumar, T., et~al.: Multi-agent pathfinding:
  Definitions, variants, and benchmarks. In: Proceedings of the International
  Symposium on Combinatorial Search. vol.~10, pp. 151--158 (2019)

\bibitem{sutton2018reinforcement}
Sutton, R.S., Barto, A.G.: Reinforcement learning: An introduction. MIT press
  (2018)

\bibitem{yakovlev2023planning}
Yakovlev, K., Andreychuk, A., Skrynnik, A., Panov, A.: Planning and learning in
  multi-agent path finding. In: Doklady Mathematics. pp.~1--6. Springer (2023)

\bibitem{ye2021mastering}
Ye, W., Liu, S., Kurutach, T., Abbeel, P., Gao, Y.: Mastering atari games with
  limited data. Advances in Neural Information Processing Systems  \textbf{34}
  (2021)

\bibitem{zerbel2019multiagent}
Zerbel, N., Yliniemi, L.: Multiagent monte carlo tree search. In: Proceedings
  of the 18th International Conference on Autonomous Agents and MultiAgent
  Systems. pp. 2309--2311 (2019)

\end{thebibliography}

\end{document}